%% file: main.tex
\documentclass[final,5p,times,twocolumn]{elsarticle}
\usepackage{framed,multirow}

\usepackage{graphicx}
\usepackage{amssymb}

\usepackage{xcolor}
\usepackage{amsmath}
\usepackage[space]{grffile}
\usepackage{fixmath}
\usepackage{array}
\usepackage{paralist}
\usepackage{color,soul}
\usepackage{hyperref}
\colorlet{usercolorname}{red!0}
\sethlcolor{usercolorname}

\setstcolor{red}

\definecolor{fuchsia}{HTML}{ca2c92}
\definecolor{gold}{HTML}{FFD700}
\definecolor{orchid}{HTML}{DA70D6}
\definecolor{cyan}{HTML}{00EEEE}
\definecolor{springgreen}{HTML}{00FF7F}
\definecolor{c1}{HTML}{41b6c4}
\definecolor{c2}{HTML}{225ea8}
\definecolor{c3}{HTML}{31a354}
\definecolor{c4}{HTML}{a1d99b}
\definecolor{c5}{HTML}{e5f5e0}

\definecolor{white}{HTML}{F5F9F7}
\definecolor{cm1}{HTML}{fef0d9}
\definecolor{cm2}{HTML}{fdcc8a}
\definecolor{cm3}{HTML}{fc8d59}
\definecolor{cm4}{HTML}{d7301f}

\definecolor{cm5}{HTML}{FABFD1}
\definecolor{cm6}{HTML}{FC718E}

\usepackage{algorithm}
\usepackage[noend]{algorithmic}

\usepackage[font=normal,labelfont=bf]{caption}
\usepackage[font=normal,labelfont=bf]{subcaption}
\usepackage{tabularx}
\usepackage{booktabs}
\usepackage{afterpage}
\usepackage{lipsum}
\usepackage{stfloats}

\usepackage{pifont}

\usepackage{footmisc}

\newcounter{daggerfootnote}

\makeatletter
\newcommand\footnoteref[1]{\protected@xdef\@thefnmark{\ref{#1}}\@footnotemark}
\makeatother

\makeatletter
\def\ps@pprintTitle{%
  \let\@oddhead\@empty
  \let\@evenhead\@empty
  \def\@oddfoot{\hfil\thepage\hfil}%
  \let\@evenfoot\@oddfoot}
\makeatother

\begin{document}


\begin{frontmatter}

\title{CURVAS-PDACVI: A Multi-Rater CT Dataset and Challenge for Pancreatic Cancer Segmentation and Vascular Involvement Assessment}

\address[sycai]{Sycai Technologies SL, Scientific and Technical Department, Barcelona, Spain}
\address[upf]{BCN Medtech, Universitat Pompeu Fabra, Barcelona, Spain}
\address[uker1]{Universit\"{a}tsklinikum Erlangen, Department of Radiology of the Uniklinikum Erlangen (UKER), Erlangen, Germany}
\address[uker2]{University Hospital Erlangen, Imaging Science Institute, Erlangen, Germany}
\address[icube]{ICUBE Laboratory, CNRS UMR-7357, University of Strasbourg, Strasbourg, France}
\address[clcc]{CLCC Institut-Strauss, Strasbourg, France}
\address[medig1]{Shenzhen Institute of Advanced Technology, Chinese Academy of Sciences, Shenzhen, China}
\address[medig2]{University of Chinese Academy of Sciences, Beijing, China}
\address[breizhseg]{Universit\'{e} de Rennes 1, CLCC Eug\`{e}ne Marquis, and INSERM UMR 1099 LTSI, Rennes, France}
\address[micdkfz]{German Cancer Research Center (DKFZ), Medical Image Computing (E230), Heidelberg, Germany}
\address[imtatlantique]{IMT Atlantique, LaTIM UMR 1101, Inserm, Brest, France}
\address[rad0]{Hospital de Matar\'{o}, Matar\'{o}, Spain}
\address[rad1]{Hospital de Sant Pau i la Santa Creu, Diagnostic Imaging Department, Barcelona, Spain}
\address[rad2]{Institut de Recerca Sant Pau - CERCA, Advanced Medical Imaging, Artificial Intelligence, and Imaging-Guided Therapy Research Group, Barcelona, Spain}
\address[icrea]{Instituci\'{o} Catalana de Recerca i Estudis Avan\c{c}ats (ICREA), Barcelona, Spain}
\address[tecnalia]{TECNALIA, Basque Research and Technology Alliance (BRTA), Bizkaia, Spain}

\author[sycai,upf]{M. Riera-Mar\'{i}n\corref{cor1}}
\ead{m.riera@sycaitechnologies.com}
\cortext[cor1]{Corresponding author}

\author[upf]{O. K. Sikha}
\author[sycai]{J. Rodr\'{i}guez-Comas}
\author[uker1,uker2]{M. S. May}
\author[icube,clcc]{T. Kirscher}
\author[clcc]{X. Coubez}
\author[icube,clcc]{P. Meyer}
\author[icube]{S. Faisan}
\author[medig1,medig2]{Z. Pan}
\author[medig1]{X. Zhou}
\author[medig1]{X. Liang}
\author[breizhseg]{C. H\'{e}mon}
\author[breizhseg]{V. Boussot}
\author[breizhseg]{J.-L. Dillenseger}
\author[breizhseg]{J.-C. Nunes}
\author[micdkfz]{K.-C. Kahl}
\author[micdkfz]{C. L\"{u}th}
\author[micdkfz]{J. Traub}
\author[imtatlantique]{P.-H. Conze}
\author[rad0]{M. M. Duh}
\author[rad1,rad2]{A. Aubanell}
\author[uker1]{R. de Figueiredo Cardoso}
\author[uker1]{S. Egger-Hackenschmidt}
\author[sycai]{J. Garc\'{i}a-L\'{o}pez}
\author[upf,icrea]{M. A. Gonz\'{a}lez-Ballester}
\author[tecnalia]{A. Galdran}

\begin{abstract}
Surgical resection remains the only potentially curative treatment for pancreatic ductal adenocarcinoma (PDAC), and eligibility depends on accurate assessment of vascular involvement (VI), i.e., tumor contact with adjacent critical vessels. Despite its importance for preoperative staging, computational VI assessment remains underexplored. Two major challenges are the lack of public datasets and the diagnostic ambiguity at the tumor-vessel interface, which leads to substantial inter-rater variability even among expert radiologists. To address these limitations, we introduce the CURVAS-PDACVI Dataset and Challenge. The dataset comprises 109 abdominal CT scans, each with five independent expert PDAC annotations (mean pairwise DSC 58.4 ± 25.7\%), a STAPLE consensus label and refined segmentations of five peripancreatic vessels. We also propose an evaluation framework that scores probabilistic predictions against all expert annotations rather than a single consensus, covering spatial overlap, multi-rater calibration, volume agreement and vessel-specific VI. Six methods submitted to the challenge were evaluated on 64 held-out test cases. Strong volumetric overlap did not translate into accurate VI assessment: the method ranked first on overlap and calibration did not rank first on VI. VI errors were largest for the superior mesenteric and portal veins, and they came mostly from vessels that at least one expert marked as involved. Calibration varied widely across methods and was not determined by multi-rater training alone: explicit uncertainty modelling or post-hoc calibration produced the best-calibrated probabilistic maps. These results indicate that volumetric overlap alone is not an adequate proxy for VI assessment, and that evaluation against multiple raters is needed to compare models on this task. The dataset is publicly available.

\end{abstract}


\end{frontmatter}

\begin{figure*}[!t]
\centering
\includegraphics[width=\linewidth]{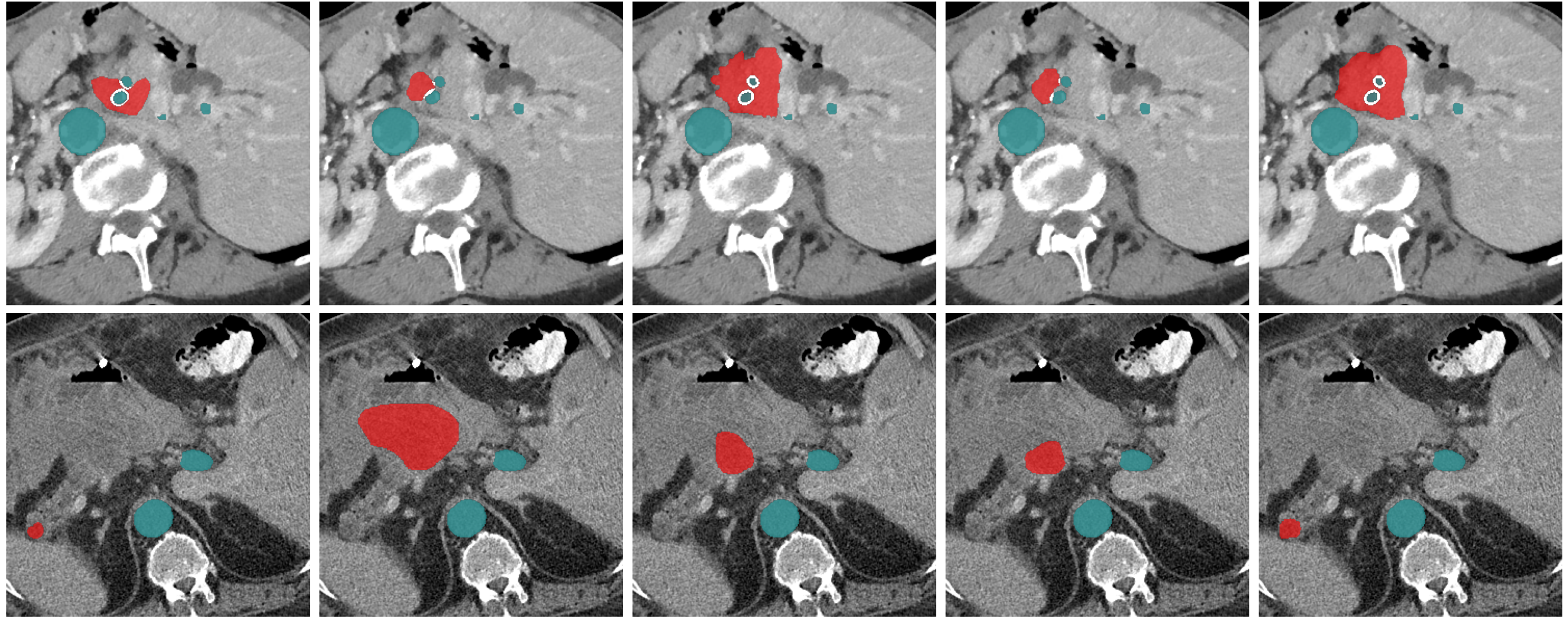}
\caption{Examples of High Diagnostic Divergence in PDAC Segmentation. 
The five columns display the independent annotations from the five human experts. 
Tumor annotations are delineated in red, adjacent vascular structures in green, and areas of vascular involvement (tumor-vessel contact) are highlighted in white. 
(\textit{Top row}) Substantial disagreement on the infiltrative borders and the specific extent of the tumor-vessel interface. 
(\textit{Bottom row}) A total lack of spatial consensus where expert annotations are scattered across entirely different anatomical regions, illustrating the uncertainty in difficult cases.
}\label{fig:examples_segmentations}
\end{figure*}

\section{Introduction}\label{sec_intro}
Precise delineation of pathological structures plays a fundamental role in the diagnosis, staging, and treatment of oncological diseases.
In particular, accurate segmentation of Pancreatic Ductal Adenocarcinoma (PDAC) is essential for assessing tumor extent, evaluating vascular involvement, and determining surgical resectability, all of which critically influence patient outcomes \cite{sun_survival_2014, brugel_vascular_2004,yao_review_2023, alidina_radiomics_2026}.
However, this task is especially challenging due to PDAC's heterogeneous appearance, ill-defined boundaries, and proximity to major vascular structures \cite{xia_detecting_2020, zhao_diagnostic_2025}.
Even experienced radiologists often disagree on PDAC boundaries and, consequently, the degree of Vascular Involvement (VI) (see Fig. \ref{fig:examples_segmentations}). 
Although recent advances in deep neural networks have achieved remarkable progress in abdominal image segmentation \cite{isensee_nnu-net_2021, bereska_artificial_2024}, the reliability and trustworthiness of the most advanced models - and thus their clinical readiness - remain limited by their inherent black-box nature \cite{jungo_effect_2018, joskowicz_inter-observer_2019}.
A notable but often neglected shortcoming in this context is the failure to model annotation variability and data uncertainty, both of which are inherent to PDAC and VI analysis \cite{buchs_vascular_2010, do_radiomics_2021}.

The aleatoric uncertainty induced by annotator variability, as well as its impact on medical image segmentation models' training and evaluation, has been addressed in many different ways.
The most widely adopted techniques attempt to reconcile discrepancies through agreement-based fusion methods like STAPLE \cite{warfield_simultaneous_2004}.
These consensus-centred strategies are mainly a systematic way of summarizing data complexity, which risks obscuring informative disagreement signal and may discard relevant information from annotator disagreement \cite{jungo_effect_2018}.
Alternative less aggressive approaches, such as random label sampling \cite{jensen_improving_2019} or variants of label smoothing \cite{sudre_lets_2019, zhang_soft_2023, islam_spatially_2021} partially mitigate this issue but still fail to capture the structure of annotator disagreement in a principled manner.
More nuanced multi-rater learning frameworks that preserve individual annotations have been shown to better capture regions of diagnostic ambiguity, thereby improving calibration and interpretability \cite{riera-marin_multi-rater_2026}. 

On the other hand, even when multiple annotations are integrated during model training, performance evaluation often relies on a single ``gold standard'' or consensus. 
This discrepancy neglects the ambiguity inherent in expert annotations for validation purposes, yielding an incomplete and potentially biased assessment of model performance. 
This limitation becomes more relevant when datasets include annotations from multiple radiologists labelling different subsets of images, as evaluations may inadvertently compare model predictions to subjective annotator-specific interpretations rather than underlying anatomy or pathology. 
As a consequence, performance metrics may not accurately reflect a model's clinical reliability~\cite{tuijn_reducing_2012}.

In clinical oncology, this variability extends beyond spatial boundaries to downstream quantitative assessments such as Vascular involvement, a primary indicator of PDAC aggressiveness and resectability \cite{brugel_vascular_2004}.
Standard clinical CT criteria for vessel involvement are known to be inherently subjective, leading to suboptimal specificity and significant inter-observer variability in clinical staging \cite{do_radiomics_2021, yang_systematic_2021, buchs_vascular_2010}.
This ambiguity poses a major challenge for automated systems, as a single deterministic prediction cannot capture the spectrum of expert interpretations that guide surgical decision-making.
Robust evaluation must therefore bridge the gap between deterministic algorithmic predictions and probabilistic clinical decision-making under uncertainty \cite{bereska_artificial_2024, shen_review_2017}.

To address these limitations, the CURVAS-PDACVI Challenge incorporates multi-rater uncertainty into PDAC segmentation and VI analysis, supported by a densely annotated dataset with five independent expert annotations per CT scan.
We also introduce an evaluation framework that goes beyond conventional overlap metrics by using probabilistic predictions to compute distribution-based measures of VI, comparing predicted involvement scores with empirical multi-rater distributions through a Wasserstein-like distance, and assessing model calibration under aleatoric uncertainty.
Together, these components provide a common framework for evaluating multi-rater calibration and VI quantification.
The following sections describe the multi-annotated dataset constructed for the competition, the methodological approaches of the CURVAS-PDACVI Challenge participants, the evaluation framework, and the benchmark results.

\section{The CURVAS-PDACVI Benchmark: Dataset}
This section describes the CURVAS-PDACVI dataset, including the annotation process, official data splits, and quality-control procedures implemented during dataset development.

\subsection{Multi-Rater Annotations in CURVAS-PDACVI}
Raw CT data for this challenge was sourced from the PANORAMA collection \cite{alves_panorama_2024}, together with initial single-rater annotations. 
A subset of 125 CT volumes was selected, ensuring no overlap with other public datasets, discarding any case with no pathologically confirmed diagnosis, and keeping only fully-manual tumor and pancreas annotations. 
After curation, data was distributed to four additional board-certified radiologists with varying levels of experience from Universitätsklinikum Erlangen, Hospital de Sant Pau, and Hospital de Mataró. 
PANORAMA original annotations were not shown to the new annotators, who were only informed about the presence of a PDAC lesion. 
Among the 125 CT scans, 16 cases were identified in which at least one radiologist did not find any PDAC to segment, resulting in exclusion from our dataset. 
Excluded and retained cases did not differ significantly in tumour volume (median 14.4 vs 18.7 ~cm$^3$, p = 0.14) or location (p = 0.14). 

In addition to re-annotating pancreas and tumoral lesions, a fifth radiologist refined the semi-automatic vascular annotations originally provided in the PANORAMA dataset. 
Specifically, the vascular structures were separated into five distinct anatomical components: the Aorta, Celiac Trunk, Porta, Superior Mesenteric Vein (SMV), and Superior Mesenteric Artery (SMA), and manual completion of partial segments was conducted. 
This is critical to enable a detailed assessment of vascular involvement (VI) for each vessel independently, allowing structure-specific quantitative evaluation. 

Eventually, participants were provided with the original CT volume and PANORAMA annotations (PDAC and pancreatic parenchyma), together with the four additional PDAC annotations, five vascular delineations, and a STAPLE consensus label of the tumoral area. 
The latter was not required for submission, but to allow participants to run a local copy of the evaluation code. 
All the annotations conforming the CURVAS-PDACVI dataset are publicly available \cite{riera_marin_2025_17552201}.

\subsection{Data Splitting}
To faithfully reflect clinical and demographic diversity of PDAC, we stratified the 109 cases across training ($N=40$), validation ($N=5$), and test ($N=64$) partitions. 
We took particular care in balancing away potential confounding variables—including patient sex, age, scanner manufacturer, tumor location, and lesion volume-across the training and test sets, as shown in Table~\ref{tab:bias_distribution}.
This ensures that observed performance differences are attributable to model behavior rather than underlying sample bias. 

\begin{table*}[!t]
\centering
\caption{Demographic, acquisition, and anatomical bias distributions across splits. 
Percentages are relative to each subset size. 
For the ``Location \& Volume" categories, values in parentheses indicate mean tumor volume (~cm$^3$, averaged over the five expert annotations) for that specific anatomical region. 
}
\label{tab:bias_distribution}
\begin{tabular}{lccccc}
\toprule
\textbf{Category} & \textbf{Variable} & \textbf{Final (N=109)} & \textbf{Train (N=40)} & \textbf{Validation (N=5)} & \textbf{Test (N=64)} \\
\midrule
\multirow{2}{*}{\textbf{Sex}} 
 & Female  & 46.79\% & 52.5\% & 80.0\% & 40.63\% \\
 & Male    & 53.21\% & 47.5\% & 20.0\% & 59.38\% \\
\midrule
\multirow{3}{*}{\textbf{Age range}} 
 & 41–60  & 21.10\% & 22.5\% & 60.0\% & 17.19\% \\
 & 61–80  & 62.39\% & 57.5\% & 40.0\% & 67.19\% \\
 & 81–100  & 16.51\% & 20.0\% & 0.0\% & 15.63\% \\
\midrule
\multirow{3}{*}{\textbf{Manufacturer}} 
 & Siemens & 26.61\% & 25.0\% & 20.0\% & 28.13\% \\
 & Philips & 42.20\% & 37.5\% & 80.0\% & 42.19\% \\
 & GE      & 8.26\% & 2.5\% & 0.0\% & 12.50\% \\
 & Canon   & 0.92\% & 2.5\% & 0.0\% & 0.0\% \\
 & TOSHIBA & 22.02\% & 32.5\% & 0.0\% & 17.19\% \\
\midrule
\multirow{3}{*}{\textbf{Location \& Volume}} 
 & Head (volume) & 58.72\% (22.91) & 65.0\% (24.01) & 60.0\% (24.16)  & 54.69\% (21.99) \\
 & Body (volume) & 27.52\% (25.45) & 17.5\% (23.43) & 20.0\% (11.93)  & 34.38\% (26.71) \\
 & Tail (volume) & 13.76\% (94.18) & 17.5\% (91.23) & 20.0\% (396.90) & 10.94\% (53.87) \\
\bottomrule
\end{tabular}
\end{table*}

\begin{figure}[!b]
\centering
\includegraphics[width=0.975\linewidth]{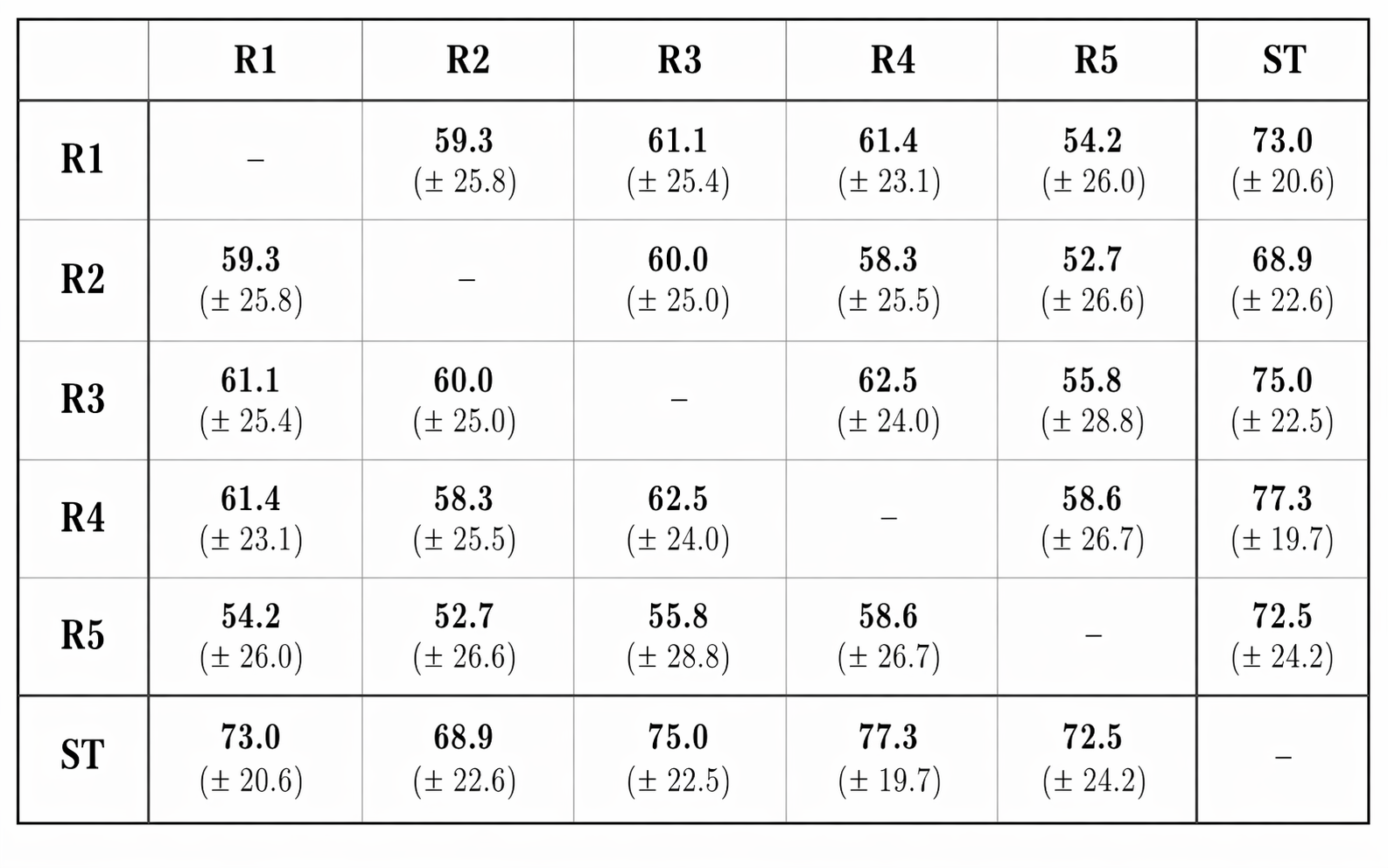}
\caption{Interrater agreement (Mean $\pm$ SD) per rater pair. 
The matrix shows average agreement across our dataset between individual experts and also the STAPLE consensus, pointing to the 1-year junior resident (Rater 5) as the primary outlier.
}\label{fig:multi}
\end{figure}

\subsection{Inter-rater Variability Analysis}
To characterize ambiguity in PDAC boundaries, we performed a statistical analysis of inter-rater agreement and disagreement patterns. 
Inter-rater variability was quantified using the Dice Similarity Coefficient (DSC) for all pairwise combinations of human annotators, as well as between each expert and the STAPLE consensus (Fig. \ref{fig:multi}). 
Human-to-human DSC scores revealed substantial aleatoric uncertainty, with a mean agreement of $58.4 \pm 25.7\%$. 

Further analysis of individual rater profiles revealed that the large standard deviation in annotations arose from a generally high agreement in many samples but a marked disagreement in a subset of outliers.
Agreement patterns were associated with clinical experience and annotation style. 
Rater 4 (3 years of experience) showed the highest average agreement with the group ($60.19\%$), followed by Rater 3 (22 years; $59.86\%$), Rater 1 (Panorama label; $59.02\%$), and Rater 2 (7 years; $57.59\%$). 
In contrast, Rater 5 (1 year) had the lowest mean agreement ($55.33\%$), consistent with the higher variability expected from early-career readers. 
These rater-specific differences highlight the bias that can arise when evaluation relies on a single annotator. 
This spatial uncertainty also has important downstream clinical implications, particularly for vascular involvement assessment, where millimetric boundary shifts may alter staging from resectable to unresectable.

\section{The CURVAS-PDACVI Benchmark: Competition}
To complement the public release of the CURVAS-PDACVI dataset, we organized an open challenge focused on uncertainty-aware PDAC segmentation and vascular involvement assessment. 
This competition was designed not only to compare algorithmic performance within a common evaluation framework, but also to examine how current segmentation methods handle multi-rater ambiguity in a clinically relevant setting.
In the following, we summarize the challenge organization, the evaluation framework, the participating methods, and the main benchmark findings.

\subsection{Organization}
The CURVAS-PDACVI challenge was hosted on the Grand Challenge platform, which provided a standardized environment for data distribution, containerized submission, and independent evaluation of participating algorithms. 
The benchmark was run over successive training, validation, and testing phases during 2025, and all submissions were assessed under the same protocol to ensure fair comparison. 
Organizer baselines were included for reference but were not eligible for prizes. 
Final results were presented during the CURVAS-PDACVI challenge session at MICCAI 2025.

\begin{table*}[!t]
\centering
\caption{Overview of submitted methods in the CURVAS-PDACVI benchmark, summarizing how each participant incorporated multi-rater supervision and uncertainty.}
\label{tab:methods}
\renewcommand{\arraystretch}{1.18}
\setlength{\tabcolsep}{4.5pt}
\small
\begin{tabularx}{\textwidth}{
>{\raggedright\arraybackslash}p{1.9cm}
>{\raggedright\arraybackslash}X
>{\raggedright\arraybackslash}X
>{\raggedright\arraybackslash}X
>{\raggedright\arraybackslash}p{2.2cm}
}
\toprule
\textbf{Method} & \textbf{Multi-rater supervision} & \textbf{Architecture / pipeline} & \textbf{Uncertainty / calibration strategy} & \textbf{External data} \\
\midrule

\textbf{TwinTrack} 
& Individual annotations not merged during training; multi-rater information used only for post-hoc calibration against mean expert agreement. 
& Two-stage nnU-Net cascade with low-resolution localization followed by high-resolution ensemble refinement in a region of interest. 
& Deep ensemble with post-hoc isotonic calibration of lesion probabilities. 
& PANORAMA Batch 4 \\

\textbf{CorpuSeg} 
& Five independent models trained separately, one per expert annotation; outputs fused at inference. 
& Two-stage nnU-Net framework with coarse localization followed by full-resolution cropped refinement. 
& Prediction-level fusion through voxel-wise averaging of expert-specific models. 
& None \\

\textbf{BreizhSeg} 
& Individual annotations preserved during training; stochastic predictions fused at inference. 
& Coarse-to-fine ResidualEncoderUNet-based pipeline. 
& Bayesian uncertainty modeling via Adaptable Bayesian Neural Network and Monte Carlo inference. 
& PANORAMA Batch 4 + MSD Pancreas \\

\textbf{MIC DKFZ} 
& Trained using all five annotations together with the STAPLE consensus as separate image--target pairs. 
& nnU-Net ResEnc-L ensemble fine-tuned on CURVAS-PDACVI. 
& Ensemble-based uncertainty with temperature scaling for calibration. 
& PANORAMA Batch 4 \\

\textbf{ROISeg} 
& Multi-rater labels fused into a single STAPLE consensus prior to training. 
& Direct full-resolution nnU-Net without explicit multi-stage refinement. 
& No explicit uncertainty modeling; conventional consensus-based segmentation. 
& None \\

\textbf{OrdSTAPLE} 
& Joint modeling of STAPLE consensus and ordinal levels of annotator agreement. 
& Two-model pipeline combining a binary STAPLE model with an ordinal agreement model. 
& Ordinal uncertainty modeling through agreement-level probabilistic prediction fusion. 
& PANORAMA Batch 4 \\

\bottomrule
\end{tabularx}
\end{table*}

\subsection{Performance Assessment and Ranking}

To provide a clinically meaningful evaluation of PDAC segmentation under multi-rater ambiguity, participant methods were assessed using four complementary metric families: segmentation quality, multi-rater calibration, probabilistic volume assessment, and vascular involvement (VI) analysis. 
This design aimed to complement conventional overlap-based benchmarking and capture both spatial accuracy and reliability under uncertainty.
Full metric definitions are provided in the Appendix.

Segmentation quality was quantified using two Dice-based measures. 
First, a conventional Dice Similarity Coefficient (DSC) was computed between each binarized prediction and a binarized STAPLE consensus mask derived from the five expert annotations. 
Second, to preserve the probabilistic nature of both predictions and multi-rater annotations, we used a threshold-averaged Dice score (Thr-DSC), computed over multiple operating points on the predicted probability map and the averaged ground-truth annotations. 
Together, these two metrics provide complementary views of performance, reflecting both standard binary segmentation quality and agreement with the full range of expert uncertainty.

To assess calibration, we used a multi-rater extension of the Expected Calibration Error (MR-ECE), computed by comparing each probabilistic prediction against each of the five expert annotations and averaging the resulting calibration errors. 
In addition, probabilistic volume estimation was evaluated using the Continuous Ranked Probability Score (CRPS), which measures how well the predicted tumor volume agrees with the distribution of volumes induced by the multiple expert annotations. 
We additionally report the predicted tumour volume of each method, from the probability map and from the binary mask, and the proportion of empty predictions (no foreground voxels) and near-empty predictions ($< 1 \textbf{ cm}^{3}$).

Finally, resectability depends on tumor-vessel contact, vascular involvement was evaluated independently for the Porta, SMV, SMA, Celiac Trunk, and Aorta. 
For each structure, predicted and reference involvement scores were represented as distributions derived from multiple thresholds and multiple annotators, respectively, and compared using a Wasserstein-based distance. 
This evaluates not only whether a model detects tumor tissue, but also whether its predicted contact with each vessel agrees with the range of expert assessments.  
To separate errors on involved and non-involved vessels, a vessel was considered involved in a case if at least one expert annotation was in contact with it, and $W_1$ was summarised separately for both groups.

Separate rankings were generated for DSC, Thr-DSC, MR-ECE, CRPS, and each of the five vessel-specific VI scores.
The final leaderboard was obtained by averaging team ranks across all metrics, with higher values indicating better performance for DSC and Thr-DSC, and lower values indicating better performance for the remaining metrics.

As a reference, we also evaluated an empty baseline, i.e. an all-zero probability map for every test case, with the same evaluation pipeline. It is reported in Tables~\ref{tab:global_performance}, \ref{tab:vascular_invasion} and~\ref{tab:supp_vi_involved} for comparison only and is not included in the ranking.

\subsection{Team Methods}

During the final testing phase, six unique algorithms were evaluated, submitted by six participating teams from international research institutions. 
Although all methods were based on variants of the nnU-Net framework, they differed substantially in how they incorporated multi-rater supervision, modeled predictive uncertainty, and balanced global segmentation quality against clinically relevant vascular precision. 
A global summary of the main design choices is provided in Table~\ref{tab:methods}.

\subsubsection*{\textbf{TwinTrack} - University of Strasbourg, France}
TwinTrack adopted a two-stage nnU-Net cascade \cite{kirscher_twintrack_2026}. 
A first low-resolution stage localized the pancreas, tumor, and surrounding vascular anatomy, while a second high-resolution stage refined the prediction within a region of interest using a deep ensemble \cite{lakshminarayanan_simple_2017}. 
Rather than integrating multiple annotations during training, TwinTrack explicitly modeled inter-rater variability through post-hoc calibration, using isotonic regression to align predicted probabilities with the mean human consensus \cite{naeini_obtaining_2015, barlow_statistical_1973}.

\subsubsection*{\textbf{CorpuSeg} - Shenzhen Institute of Advanced Technology}
CorpuSeg also followed a two-stage nnU-Net strategy, but handled multi-rater supervision differently. 
After coarse tumor localization, five independent second-stage models were trained, one per expert annotation, and their probabilistic outputs were averaged at inference time. 
This design allowed the final prediction to reflect annotator variability through prediction-level fusion.

\subsubsection*{\textbf{BreizhSeg} - University of Rennes 1, CLCC Eugène Marquis, INSERM UMR LTSI}
BreizhSeg combined coarse-to-fine segmentation with Bayesian uncertainty modeling. 
The method built on a pretrained ResidualEncoderUNet and converted it into an Adaptable Bayesian Neural Network \cite{franchi_make_2024} by introducing stochastic perturbations into normalization layers. 
Multiple stochastic forward passes were then used at inference time to approximate the posterior predictive distribution \cite{hemon_towards_2025}, and the resulting probabilistic segmentations were fused into a final consensus prediction.

\subsubsection*{\textbf{MIC DKFZ} - German Cancer Research Center}
MIC DKFZ used a pretrained nnU-Net ResEnc-L ensemble, subsequently fine-tuned on the CURVAS-PDACVI training set using all available annotations, including the STAPLE consensus. 
Uncertainty was estimated from ensemble softmax outputs and further calibrated through temperature scaling. 
This approach emphasized strong baseline segmentation performance while explicitly addressing calibration at inference time.

\subsubsection*{\textbf{ROISeg} - Chinese Academy of Sciences}
ROISeg represented the most direct training strategy among submissions. Instead of preserving the individual annotations, the five expert labels were first fused into a single STAPLE consensus, which was then used to train a full-resolution nnU-Net. 
This simplified the learning problem to conventional consensus-based segmentation, without explicit multi-rater or uncertainty modeling.

\subsubsection*{\textbf{OrdSTAPLE} - Universitat Pompeu Fabra, Sycai Medical}
OrdSTAPLE, submitted by the organizers as a reference baseline, explicitly modeled rater disagreement through an ordinal formulation. 
The method combined a standard STAPLE-trained binary model with a second model trained to predict increasing levels of annotator agreement using an ordinal objective. 
Their probabilistic outputs were then merged to balance the sharpness of conventional binary segmentation with the more diffuse behavior of ordinal uncertainty modeling.

\subsection{Challenge Results}

\subsubsection{Quantitative Benchmark Results}
\begin{table*}[!t]
\centering
\caption{Global benchmark performance on the test set. Values denote the mean across cases, with overall metric rank shown in parentheses. Higher values indicate better performance for DSC and Thr-DSC, whereas lower values indicate better performance for MR-ECE and CRPS.}
\label{tab:global_performance}
\renewcommand{\arraystretch}{1.35}
\setlength{\tabcolsep}{22pt}

\begin{tabular}{lcccc}
\toprule
\textbf{Method} & \textbf{DSC (\%)} & \textbf{Thr-DSC (\%)} & \textbf{MR-ECE ($\times 10^{-3}$)} & \textbf{CRPS (cm$^3$)} \\
\midrule
\textbf{TwinTrack}  & 55.76 (5)          & 56.93 (4)          & 29.6 (2)          & 5.924 (2)          \\
\textbf{CorpuSeg}   & 58.94 (4)          & 58.01 (3)          & 30.5 (3)          & 10.792 (6)         \\
\textbf{BreizhSeg}  & \textbf{71.04 (1)} & \textbf{64.01 (1)} & \textbf{25.7 (1)} & 7.320 (4)          \\
\textbf{MIC DKFZ}   & 66.21 (2)          & 59.57 (2)          & 32.2 (4)          & 7.256 (3)          \\
\textbf{ROISeg}     & 59.28 (3)          & 55.32 (5)          & 34.5 (5)          & \textbf{5.352 (1)} \\
\textbf{OrdSTAPLE}  & 54.05 (6)          & 48.73 (6)          & 40.3 (6)          & 8.592 (5)          \\
\textbf{Empty mask}  & 0.00 (--)          & 2.86 (--)          & 58.9 (--)          & 13.110 (--)          \\
\bottomrule
\end{tabular}
\end{table*}
\begin{table*}[!b]
\centering
\caption{Vessel-specific vascular invasion assessment on the test set. Values correspond to the mean Wasserstein distance ($W_1$, in degrees) between predicted and reference invasion distributions for each vascular structure, with rank shown in parentheses. Lower values indicate better agreement with expert-derived vascular invasion profiles.}
\label{tab:vascular_invasion}
\renewcommand{\arraystretch}{1.35}
\setlength{\tabcolsep}{20pt}
\begin{tabular}{lccccc}
\toprule
\textbf{Method} & \textbf{PORTA} & \textbf{AORTA} & \textbf{SMA} & \textbf{SMV} & \textbf{CELIAC TRUNK} \\
\midrule
\textbf{TwinTrack}  & \textbf{29.03 (1)} & \textbf{6.08 (1)}  & 28.69 (3)          & \textbf{34.03 (1)} & \textbf{14.48 (1)} \\
\textbf{CorpuSeg}   & 29.09 (2)          & 7.13 (2)           & \textbf{28.08 (1)} & 40.06 (3)          & 14.80 (2)          \\
\textbf{BreizhSeg}  & 35.94 (5)          & 9.12 (4)           & 29.34 (4)          & 43.34 (4)          & 22.38 (4)          \\
\textbf{MIC DKFZ}   & 33.53 (4)          & 10.76 (5)          & 28.50 (2)          & 35.25 (2)          & 22.48 (5)          \\
\textbf{ROISeg}     & 33.16 (3)          & 7.91 (3)           & 41.91 (6)          & 45.25 (6)          & 22.80 (6)          \\
\textbf{OrdSTAPLE}  & 41.81 (6)          & 13.51 (6)          & 33.66 (5)          & 44.16 (5)          & 19.53 (3)          \\
\textbf{Empty mask}  & 67.16 (--)          & 13.79 (--)          & 63.37 (--)          & 95.93 (--)          & 48.09 (--)          \\
\bottomrule
\end{tabular}
\end{table*}

We first evaluated all submitted methods using the full benchmark metric set, covering both global segmentation quality and probabilistic reliability. 
Table~\ref{tab:global_performance} summarizes the mean test-set results for DSC, Thr-DSC, MR-ECE, and CRPS. 
Across methods, global segmentation performance was competitive, with the strongest approaches reaching mean DSC values above 65\% and Thr-DSC values close to 60-65\%. 
The BreizhSeg team obtained the best overall overlap and calibration profile, ranking first in DSC, Thr-DSC, and MR-ECE, whereas team ROISeg achieved the lowest CRPS. 
All methods outperformed the empty baseline on every global metric. The baseline's non-zero Thr-DSC (2.86\%) arises because thresholds at which neither the prediction nor the averaged annotations contain tumour are scored as perfect agreement.

Table~\ref{tab:vascular_invasion} reports the vessel-specific vascular involvement (VI) errors for the Aorta, Porta, SMV, SMA, and Celiac Trunk. 
This analysis reveals a different ranking profile from that observed in the global metrics. 
Team TwinTrack achieved the lowest $W_1$ error for four of the five vascular structures, while team CorpuSeg obtained the best result for the SMA and remained consistently competitive across the remaining vessels. 
Across all methods, the Aorta was the least challenging structure to assess, whereas the SMV and the portal vein yielded the highest errors, suggesting that venous involvement is harder to assess. 
For reference, the empty baseline yields $W_1$ between 13.79 (Aorta) and 95.93 (SMV), roughly 1.5–3 times the error of the participating methods. The exception is the Aorta, where OrdSTAPLE (13.51) and MIC DKFZ (10.76) perform close to the baseline (13.79).

Median predicted volumes (10.5–20.0 ~cm$^3$) were close to the expert median (17.9 ~cm$^3$). Although OrdSTAPLE ranked lowest on global overlap and calibration, its binary volume distribution was the closest to the expert distribution at cohort level (median 18.4 vs 17.9 cm³; Table~\ref{tab:supp_volumes}), and its VI error on involved vessels was similar to that of MIC DKFZ and BreizhSeg (51.7° vs 51.5–51.9°; Table~\ref{tab:supp_vi_involved}), the two top-ranked methods on global overlap. Its lower VI rank was driven partly by false-positive contact on non-involved vessels, consistent with the more diffuse predictions of ordinal agreement modelling. However, TwinTrack returned an empty mask in 18.8\% of test cases (12/64), compared with 1.6–7.8\% for the other methods, whereas no expert annotation was empty (Table~\ref{tab:supp_volumes}).

VI errors were concentrated in vessels marked as involved by at least one expert (mean $W_1$ 42–60° across methods) and were small where no expert marked involvement (1–10°). False-positive contact was highest for OrdSTAPLE (9.9°) and BreizhSeg (4.8°) (Table~\ref{tab:supp_vi_involved}).

Taken together, these results show that strong global segmentation performance does not necessarily translate into accurate assessment of localized vascular involvement. 
The proposed benchmark therefore distinguished between methods that excel in volumetric tumor delineation and those that remain more reliable at the anatomically critical boundaries that drive PDAC resectability assessment.

\subsubsection{Statistical Robustness of the Benchmark Ranking}
\begin{figure*}[!t]
    \centering
    \includegraphics[width=\textwidth]
    {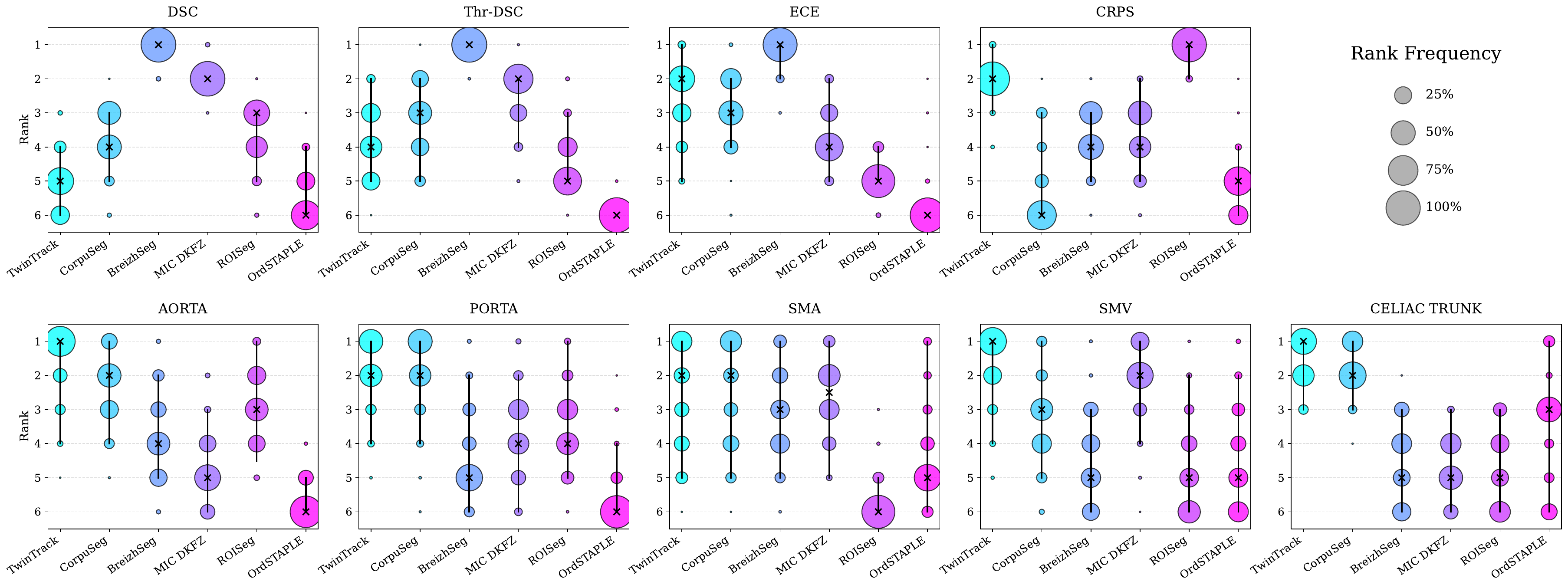}
    \caption{Ranking stability via bootstrap analysis. Bubble charts illustrating the rank frequency (1st to 6th) achieved by each participating team across 500 bootstrap iterations for all nine metrics. The size of each round marker corresponds to the percentage of iterations in which a team achieved a given rank.}
    \label{fig:bootstrap_ranking}
\end{figure*}
To assess whether our benchmark ranking was robust to the specific composition of the test cohort, we performed a bootstrap analysis with 500 resampling iterations, see Fig.~\ref{fig:bootstrap_ranking}. 
Across global metrics, the highest-ranked methods showed stable ranking profiles: team BreizhSeg consistently occupied the top positions for DSC, Thr-DSC, and MR-ECE, while TwinTrack and CorpuSeg remained the most competitive methods for vascular involvement assessment. 
These results indicate that the main ranking trends were not driven by isolated test cases, but reflected consistent performance differences across repeated cohort resampling.

We further evaluated pairwise statistical significance using Wilcoxon signed-rank tests on a case-by-case basis. 
For global metrics, several differences remained significant, particularly for calibration and overlap. 
In contrast, most pairwise comparisons for the vessel-specific vascular involvement metrics did not reach statistical significance. 
This apparent discrepancy between stable average rankings and weaker pairwise significance is consistent with a subset of difficult cases that degrade all methods and increase case-level variance.

Taken together, these analyses support the robustness of the benchmark ranking at the population level, while also showing that fine-grained vascular assessment remains strongly limited by case difficulty and inter-rater ambiguity.

\section{Discussion}

\begin{figure*}[!t]
    \centering
    \includegraphics[width=\textwidth]{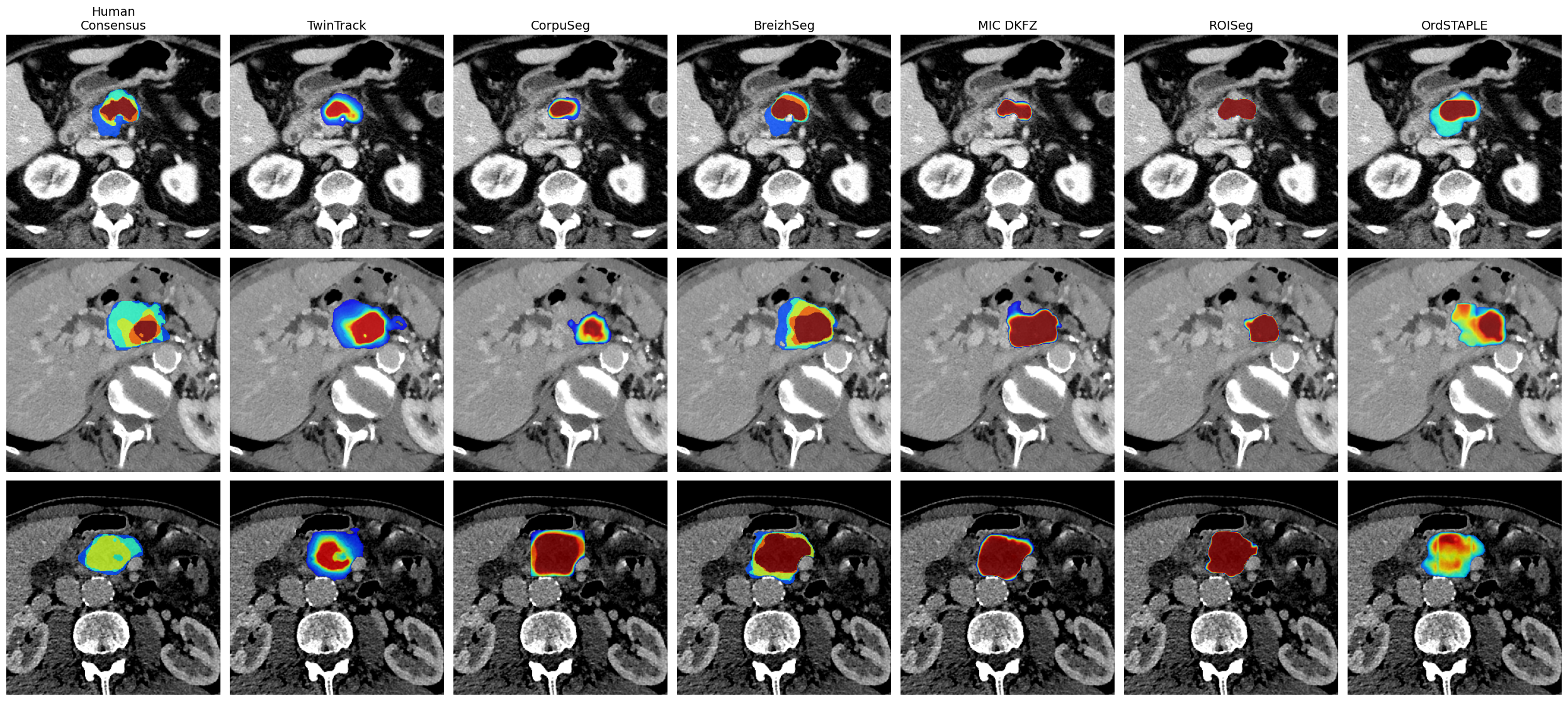}
\caption{Representative predictions in three test cases with increasing (top to bottom) inter-rater disagreement. 
First column: CT slice with the expert agreement map. 
Second-to-last columns: predictions of different participants, highlighting different confidence behaviors at the tumor-vessel interface. 
Methods differ in how graded their outputs are: ROISeg and MIC DKFZ produce sharp, near-binary boundaries, whereas OrdSTAPLE, TwinTrack and BreizhSeg produce graded probability maps.}
    \label{fig:clinical_examples}
\end{figure*}

\subsection{Benchmark findings}
On the full test cohort, several methods achieved similar global overlap and calibration scores but ranked differently on vessel-specific VI. Strong volumetric agreement therefore did not guarantee equally accurate VI assessment, particularly for the venous structures, which were the most challenging across methods.

A second finding is that the way methods handled multi-rater supervision produced different failure modes rather than a uniform ranking. BreizhSeg and MIC DKFZ performed strongly under global overlap metrics, whereas TwinTrack and CorpuSeg were more competitive at the vascular interface. Failure modes also differed. TwinTrack returned an empty mask in 18.8\% of test cases, whereas OrdSTAPLE produced more diffuse probability maps (median probabilistic volume 25.8 vs 17.9 ~cm$^3$ for the experts) while keeping binary volumes close to the expert distribution, at the cost of false-positive vessel contact. Finally, multi-rater training alone did not determine calibration: the best-calibrated maps came from explicit Bayesian uncertainty modelling (BreizhSeg) or post-hoc calibration (TwinTrack).

Overall, these results indicate that the main challenge in VI-focused PDAC segmentation is not simply recovering tumor volume, but doing so in a way that remains calibrated and anatomically reliable at clinically critical boundaries. 
Our benchmark supports evaluating methods not only through aggregate overlap, but also through uncertainty-aware, multi-metric frameworks that better reflect downstream clinical goals.

\subsection{Clinical implications}
Our findings suggest that global volumetric overlap alone is not an adequate proxy for clinical usefulness in PDAC staging. 
In practice, surgical decision-making depends critically on the local characterization of the tumor-vessel interface, particularly around structures such as the SMV, PORTA, and SMA, where even small boundary differences may alter resectability assessment. 
Our quantitative results show that models with strong average segmentation scores do not necessarily preserve the same advantage at the tumor–vessel interface.

This distinction becomes particularly relevant when considering predictive confidence. 
In ambiguous cases, several binary-target methods tended to produce sharp and visually confident boundaries even when expert agreement was weak, whereas disagreement-aware approaches more often generated diffuse probability maps that visually better reflected the uncertainty present in the annotations, see Fig.~\ref{fig:clinical_examples}. 
From a clinical perspective, this difference is meaningful: an overconfident false-positive vascular contact may incorrectly suggest unresectability, whereas an overconfident false-negative prediction may underestimate local involvement. 
In both scenarios, poorly calibrated confidence can be as clinically relevant as geometric error itself. 
Both failure modes were observed in our results: empty predictions risk missing vessel contact altogether, whereas the false-positive contact of OrdSTAPLE on non-involved vessels risks overstating it. Which error is preferable depends on the clinical use, since a missed and a false contact carry different consequences for resectability assessment.

These observations reinforce the idea that probabilistic segmentation should be understood not only as a more complete model output, but also as a mechanism to communicate confidence and support downstream review in cases where both the imaging evidence and the expert labels remain uncertain.

\subsection{Limitations}
Several limitations should be considered when interpreting our benchmark results. 
First, although the dataset is densely annotated by multiple experts, the test cohort remains relatively small (64 cases). 
This limits the statistical power of pairwise comparisons, and some ranking differences, particularly for vessel-specific metrics, should be interpreted with caution. 
Moreover, the VI metric cannot penalize under-segmentation on vessels that no expert marked as involved, where an empty prediction scores W$_1 = 0$ (Table~\ref{tab:supp_vi_involved}), so VI scores partly reward conservative predictions when contact is small or uncertain.

Second, the benchmark ground truth is derived from expert radiological annotations rather than from surgical or pathological mapping of the true tumor extent and vascular involvement. 
While this reflects the reality of preoperative imaging practice, it also means that the reference standard remains conditional on expert interpretation and therefore inherits its uncertainty and potential bias.

Finally, the participating methods showed limited architectural diversity, with all submissions relying on variants of the nnU-Net framework. 
As a result, the benchmark primarily captures differences in training strategy, uncertainty modeling, and annotation handling, rather than the broader design space of modern segmentation architectures. 
Future editions should therefore expand both cohort size and methodological diversity, while also exploring evaluation protocols that can better link radiological ambiguity to downstream clinical decision-making.

\section{Conclusion}
This paper presents the CURVAS-PDACVI dataset and challenge: a publicly released multi-rater CT dataset for PDAC segmentation, together with an open challenge and evaluation framework for vascular involvement assessment under inter-rater uncertainty. 
Our results show that strong global segmentation performance does not necessarily translate into accurate characterization of the tumor–vessel interface.  
Methods also differed in their failure modes, from missed tumors to false-positive vessel contact, which a single overlap score does not reveal. 
Future work should expand the size and diversity of the cohort, include cases with surgical or pathological confirmation of vascular involvement, and evaluate whether calibrated probabilistic outputs improve radiologists' assessment of resectability.

\section*{CRediT authorship contribution statement}

\textbf{Meritxell Riera-Marín:} Conceptualization, Methodology, Software, Investigation, Writing - Original Draft, Writing - Review \& Editing, Data Curation, Formal analysis, Validation.
\textbf{Sikha O K:} Conceptualization, Software, Visualization. 
\textbf{Júlia Rodríguez-Comas:} Conceptualization, Funding acquisition. 
\textbf{Matthias Stefan May:} Conceptualization, Data Curation, Funding acquisition. 
\textbf{Pierre-Henri Conze: } Conceptualization, Writing - Review \& Editing. 
\textbf{Anton Aubanell:} Data Curation.
\textbf{Maria Montserrat Duh:} Data Curation.
\textbf{Rubén de Figueiredo Cardoso:} Data Curation.
\textbf{Saskia Egger-Hackenschmidt:} Data Curation.
\textbf{Javier García-López:} Supervision, Writing - Review \& Editing, Funding acquisition. 
\textbf{Miguel A. González-Ballester:} Supervision, Writing - Review \& Editing, Funding acquisition. 
\textbf{Adrian Galdran:} Conceptualization, Supervision, Writing - Review \& Editing, Funding acquisition. 
\textbf{Tristan Kirscher:} Software, Investigation, Writing - Review \& Editing.
\textbf{Xavier Coubez:} Software, Investigation, Writing - Review \& Editing. 
\textbf{Philippe Meyer:} Software, Investigation.
\textbf{Sylvain Faisan:} Software, Investigation.
\textbf{Zhaohong Pan:} Software, Investigation, Writing - Review \& Editing.
\textbf{Xiang Zhou:} Software, Investigation.
\textbf{Xiaokun Liang:} Software, Investigation.
\textbf{Cédric Hémon:} Software, Investigation, Writing - Review \& Editing.
\textbf{Valentin Boussot:} Software, Investigation, Writing - Review \& Editing.
\textbf{Jean-Louis Dillenseger:} Software, Investigation.
\textbf{Jean-Claude Nunes:} Software, Investigation.
\textbf{Kim-Céline Kahl:} Software, Investigation, Writing - Review \& Editing. 
\textbf{Carsten Lüth:} Software, Investigation. 
\textbf{Jeremias Traub:} Software, Investigation. 

\section*{Code availability}
To support users with the evaluation of the algorithms, we have made the evaluation code used in this manuscript publicly available in the Challenge's GitHub repository \cite{curvas_github}.

\section*{Declaration of Competing Interest}
Financial support received by the authors is listed in the Acknowledgments. M.R.-M., J.R.-C. and J.G.-L. are affiliated with Sycai Technologies SL, which co-funded this work through an industrial doctorate. The other authors declare no competing interests.

\section*{Data availability}
All data has been uploaded on Zenodo \cite{riera_marin_2025_17552201}.

\section*{Declaration of generative AI and AI-assisted technologies in the manuscript preparation process}

During the preparation of this work, the author(s) used Claude (Anthropic) to improve the language of the manuscript. The author(s) reviewed and edited the output as needed and take full responsibility for the content of the published article.

\section*{Acknowledgments}
This work was supported by the Catalan Government through the "Doctorats Industrials" program, specifically the industrial doctorate AGAUR 2021-063, in collaboration with Sycai Technologies SL. 
The challenge is part of the Proyectos de Colaboración Público-Privada (CPP2021-008364), funded by MCIN/AEI/10.13039/501100011033 and co-financed by the European Union through the NextGenerationEU/PRTR initiative. 
Additional funding was provided by the "NUM 2.0" project (FKZ: 01KX2121) and the Eureka Eurostars-3 joint program (1661 DARE-KPL), co-funded by the Horizon Europe Research and Innovation Framework Programme of the European Union (FKZ: 347 01QE2249C).
A.G. is supported by grant RYC2022-037144-I, funded by MCIN/AEI/10.13039/501100011033 and co-financed by FSE+.
The work of T.K. is supported by the Interdisciplinary Thematic Institute HealthTech, as part of the ITI 2021-2028 program of the University of Strasbourg, CNRS and Inserm, partially supported by IdEx Unistra (ANR10-IDEX-0002) and SFRI (STRAT’US project, ANR-20-SFRI-0012) under the framework of the French Investments for the Future Program.
The work of Z.P., X.Z. and X.L. is supported by grants from the National Key Research and Development Program of China (2023YFC2411502) and the National Natural Science Foundation of China (82202954).
V.B. was supported by the Brittany Region through its Allocations de Recherche Doctorale framework and by the French National Research Agency as part of the VATSop project (ANR-20-CE19-0015). 
C.H. was supported through a PhD scholarship grant from Elekta AB. 

\newpage
\appendix
\setcounter{figure}{0}
\setcounter{table}{0}

\section{Supplementary details}

\subsection{Dataset}

The original PANORAMA CT scans used in this work are available at \url{https://zenodo.org/records/11034178}, and the CURVAS-PDACVI benchmark release, including the multi-rater annotations and evaluation resources, is available at \url{https://zenodo.org/records/17552201}. 
In the released repository, each case follows a standardized file organization: \texttt{annotation\_1.nii.gz} contains the original PANORAMA segmentation, including PDAC (\texttt{label=1}) and pancreatic parenchyma (\texttt{label=2}); \texttt{annotation\_X.nii.gz}, with \texttt{X=2,\dots,5}, contains the additional PDAC annotations provided by the expert raters; \texttt{annotation\_vascular.nii.gz} contains the five vascular structures considered in the benchmark, namely Porta (\texttt{1}), SMV (\texttt{2}), Aorta (\texttt{3}), Celiac Trunk (\texttt{4}), and SMA (\texttt{5}); \texttt{annotation\_staple.nii.gz} contains the STAPLE consensus used for binary evaluation; and \texttt{image.nii.gz} corresponds to the CT volume.

For completeness, Table~\ref{tab:dataset_bias_extended} reports the full demographic, acquisition, and anatomical distributions before and after curation, including the initial 125-case subset drawn from PANORAMA and the final 109-case cohort used in the benchmark.
The additional columns are included to document the effect of the curation step, in which 16 studies were excluded because at least one expert did not identify a PDAC lesion, and to show that the final train/validation/test splits remained broadly representative of the original selected cohort.

\subsection{Challenge Organization}
The benchmark was hosted on the Grand Challenge platform at \url{https://curvas-pdacvi.grand-challenge.org/}. 
The challenge proceeded through a staged protocol comprising a training release (May 2025), an open sanity-check phase (June 2025), an open validation phase (June–July 2025), and a closed testing phase (July–August 2025), followed by organizer-side result analysis in August–September 2025 and public announcement of winners at MICCAI 2025. 
Due to time and memory limitations encountered by participants during testing, the testing phase was extended. The workshop took place on September 27th, 2025, at the Daejeon Convention Center, South Korea, and included oral presentations by the top-performing teams. 
According to the benchmark statistics page, the challenge involved 62 registered participants, with 125 submissions in the sanity-check phase, 55 in validation, and 32 in testing.

\begin{table*}[t]
\centering
\caption{Demographic, acquisition, and anatomical bias distributions across the different dataset splits. Percentages are expressed relative to each subset size. For the ``Location \& Volume'' categories, the values in parentheses indicate the mean tumor volume in ~cm$^3$ for that specific anatomical region (averaged over the five expert annotations). Mean volumes are not reported for the full collection (N=125), as the excluded studies are not part of the curated reference.}
\label{tab:dataset_bias_extended}
\renewcommand{\arraystretch}{1.15}
\setlength{\tabcolsep}{4.5pt}
\small
\begin{tabularx}{\textwidth}{
l
l
*{5}{>{\centering\arraybackslash}X}
}
\toprule
\textbf{Category} & \textbf{Variable} & \textbf{All (N=125)} & \textbf{Final (N=109)} & \textbf{Train (N=40)} & \textbf{Validation (N=5)} & \textbf{Test (N=64)} \\
\midrule

\multirow{2}{*}{Sex}
& Female & 48.8\% & 46.79\% & 52.5\% & 80.0\% & 40.63\% \\
& Male   & 51.2\% & 53.21\% & 47.5\% & 20.0\% & 59.38\% \\
\midrule

\multirow{3}{*}{Age range}
& 41--60  & 20.8\% & 21.10\% & 22.5\% & 60.0\% & 17.19\% \\
& 61--80  & 61.6\% & 62.39\% & 57.5\% & 40.0\% & 67.19\% \\
& 81--100 & 17.6\% & 16.51\% & 20.0\% & 0.0\%  & 15.63\% \\
\midrule

\multirow{5}{*}{Manufacturer}
& Siemens               & 28.0\%  & 26.61\% & 25.0\% & 20.0\% & 28.13\% \\
& Philips               & 43.2\%  & 42.20\% & 37.5\% & 80.0\% & 42.19\% \\
& GE Medical Systems    & 7.20\%  & 8.26\%  & 2.5\%  & 0.0\%  & 12.50\% \\
& Canon Medical Systems & 0.80\%  & 0.92\%  & 2.5\%  & 0.0\%  & 0.0\% \\
& TOSHIBA               & 20.80\% & 22.02\% & 32.5\% & 0.0\%  & 17.19\% \\
\midrule

\multirow{3}{*}{Location \& Volume}
& Head (volume) & 60.0\% & 58.72\% (22.91) & 65.0\% (24.01) & 60.0\% (24.16) & 54.69\% (21.99) \\
& Body (volume) & 24.8\% & 27.52\% (25.45) & 17.5\% (23.43) & 20.0\% (11.93) & 34.38\% (26.71) \\
& Tail (volume) & 15.2\% & 13.76\% (94.18) & 17.5\% (91.23) & 20.0\% (396.90) & 10.94\% (53.87) \\
\bottomrule
\end{tabularx}
\end{table*}

\subsection{Performance Assessment}

Participants were required to submit, for each study, a binarized PDAC segmentation together with the corresponding probabilistic PDAC map. 
Vascular structures were provided for local testing and evaluation only, and did not need to be segmented by participants.
Below we cover the formulation of our performance evaluation framework\footnote{The official evaluation code is publicly available at
\url{https://github.com/SYCAI-Technologies/curvas-challenge/tree/main/CURVAS-PDACVI_2025/evaluation_metrics}.}

\paragraph{Segmentation quality}
Let $\hat{y}\in\{0,1\}^{V}$ denote the binarized prediction, let $p\in[0,1]^V$ denote the probabilistic prediction, let $a^{(k)}\in\{0,1\}^{V}$, $k=1,\dots,5$, denote the five expert annotations, and let $s\in\{0,1\}^{V}$ denote the binarized STAPLE consensus. We define the standard Dice score as:
\[
\mathrm{DSC}(\hat{y},s)=
\frac{2\sum_{i=1}^{V}\hat{y}_i s_i}
{\sum_{i=1}^{V}\hat{y}_i+\sum_{i=1}^{V}s_i+\varepsilon}.
\]

To retain multi-rater information, we also compute a threshold-averaged Dice score. First, the averaged annotation is:
\[
\bar{a}_i=\frac{1}{5}\sum_{k=1}^{5} a^{(k)}_i.
\]
Using thresholds:
\[
\mathcal{T}=\{0.10,\,0.24,\,0.38,\,0.52,\,0.66,\,0.80\},
\]
we define:
\[
\mathrm{Thr\mbox{-}DSC}
=
\frac{1}{|\mathcal{T}|}\sum_{t\in\mathcal{T}}
\mathrm{DSC}\!\left(\mathbb{I}[p>t],\,\mathbb{I}[\bar{a}>t]\right),
\]
with the convention that the Dice score is set to $1$ when both thresholded masks are empty.

\paragraph{Multi-rater calibration}
Calibration is assessed against each annotator separately and then averaged. Let $B$ denote the cropped evaluation region, defined as the union bounding box of the five annotations with fixed padding. For voxel $i\in B$, the foreground probability is $p_i$ and the corresponding two-class probability vector is:
\[
q_i=(1-p_i,\;p_i).
\]
After binning voxel confidences into $M=50$ bins, the Expected Calibration Error for annotator $k$ is:
\[
\mathrm{ECE}^{(k)}
=
\sum_{m=1}^{M}
\frac{|B_m^{(k)}|}{|B|}
\left|
\mathrm{acc}\!\left(B_m^{(k)}\right)
-
\mathrm{conf}\!\left(B_m^{(k)}\right)
\right|,
\]
where $B_m^{(k)}$ is the set of voxels assigned to bin $m$, and
$\mathrm{acc}(B_m^{(k)})$ and $\mathrm{conf}(B_m^{(k)})$ denote the empirical accuracy and average confidence in that bin, respectively. The final multi-rater calibration score is:
\[
\mathrm{MR\mbox{-}ECE}
=
\frac{1}{5}\sum_{k=1}^{5}\mathrm{ECE}^{(k)}.
\]

\paragraph{Probabilistic volume assessment}
For each annotator $k$, let:
\[
v^{(k)}=\Delta_v\sum_{i=1}^{V}\mathbb{I}[a_i^{(k)}=1]
\]
be the annotated tumor volume, where $\Delta_v$ is the voxel volume after any resampling correction. The predicted probabilistic volume is:
\[
\hat{v}=\Delta_v\sum_{i=1}^{V} p_i.
\]
From the five expert volumes, we compute:
\[
\mu_v=\frac{1}{5}\sum_{k=1}^{5}v^{(k)},
\qquad
\sigma_v=
\sqrt{\frac{1}{5}\sum_{k=1}^{5}\left(v^{(k)}-\mu_v\right)^2}.
\]
The reference cumulative distribution is obtained from the Gaussian model
$\mathcal{N}(\mu_v,\sigma_v^2)$ and evaluated on a finite grid:
\[
x_1,\dots,x_L,\qquad L=100,
\]
uniformly spanning the interval between the 1st and 99th percentiles of that Gaussian. Let:
\[
F_\ell = \Phi\!\left(\frac{x_\ell-\mu_v}{\sigma_v}\right),
\qquad
H_\ell(\hat{v})=\mathbb{I}[x_\ell\geq \hat{v}],
\]
and let $\Delta_x=x_{\ell+1}-x_\ell$ denote the grid spacing. We then use the following discrete version of the CRPS:
\[
\mathrm{CRPS}
=
\Delta_x\sum_{\ell=1}^{L}
\left(F_\ell-H_\ell(\hat{v})\right)^2.
\]

\paragraph{Vascular involvement assessment}
Vascular involvement is evaluated independently for the five relevant structures:
PORTA, SMV, SMA, CELIAC TRUNK, and AORTA. For each vessel $s$ and plane
$\pi\in\{\mathrm{cor},\mathrm{sag},\mathrm{ax}\}$, a slice-wise contact angle is computed and the maximum angle over slices is retained. Denoting by
$g_{k,s,\pi}$ the resulting angle for annotator $k$, and by $\hat{g}_{j,s,\pi}$ the corresponding angle obtained from the thresholded prediction at threshold $t_j\in\mathcal{T}$, we define:
\[
\mu^{\mathrm{gt}}_{s,\pi}=\frac{1}{5}\sum_{k=1}^{5}g_{k,s,\pi},
\qquad
\sigma^{\mathrm{gt}}_{s,\pi}=
\sqrt{\frac{1}{5}\sum_{k=1}^{5}\left(g_{k,s,\pi}-\mu^{\mathrm{gt}}_{s,\pi}\right)^2},
\]
and:
\[
\mu^{\mathrm{pred}}_{s,\pi}=\frac{1}{|\mathcal{T}|}\sum_{j=1}^{|\mathcal{T}|}\hat{g}_{j,s,\pi},
\qquad
\sigma^{\mathrm{pred}}_{s,\pi}=
\sqrt{\frac{1}{|\mathcal{T}|}\sum_{j=1}^{|\mathcal{T}|}\left(\hat{g}_{j,s,\pi}-\mu^{\mathrm{pred}}_{s,\pi}\right)^2}.
\]

For each $(s,\pi)$ pair, ground-truth and prediction are converted into Gaussian distributions with a small smoothing constant added to the standard deviation. Their densities are then sampled on a discrete grid:
\[
z_1,\dots,z_N,\qquad N=1000,\qquad z_n\in[0,360].
\]
Let $p_n$ and $q_n$ denote the resulting non-negative normalized weights after sanitization (replacement of NaN/Inf by zero, clipping negative values to zero, and normalization to unit mass). Writing the cumulative sums:
\[
P_n=\sum_{r=1}^{n}p_r,
\qquad
Q_n=\sum_{r=1}^{n}q_r,
\]
the plane-wise vascular involvement discrepancy is computed as the discrete 1D Wasserstein distance:
\[
W_{s,\pi}
=
\Delta_z\sum_{n=1}^{N}\left|P_n-Q_n\right|,
\]
where $\Delta_z=z_{n+1}-z_n$.

Fallback rules are applied when one or both sampled distributions are effectively empty. If both are empty, the score is set to $0$. If only one is empty and the non-empty distribution is degenerate, the score is set to the absolute difference between its mean and $0$; otherwise, a penalty of $360$ is assigned.


\paragraph{Final ranking}
Separate rankings are computed for DSC, Thr-DSC, MR-ECE, CRPS, and the five vessel-specific vascular involvement scores. DSC and Thr-DSC are ranked in descending order, while MR-ECE, CRPS, and all vascular involvement scores are ranked in ascending order. The final leaderboard is obtained by averaging the resulting ranks across the nine evaluation axes.

\begin{figure*}[!t]
    \centering
    \includegraphics[width=0.85\textwidth]{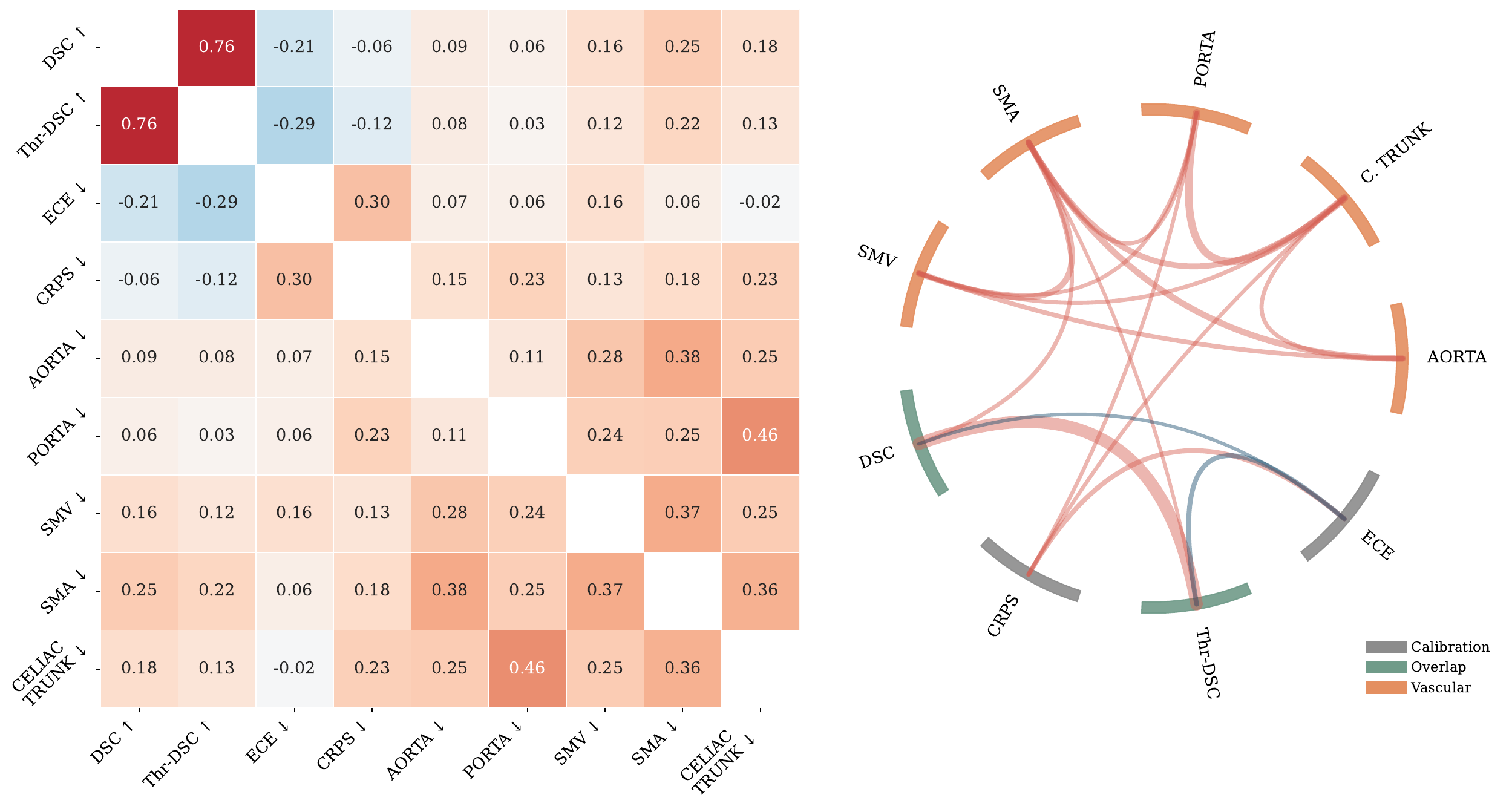}    
    \caption{Performance metric inter-correlation analysis. Overlap metrics were strongly correlated with each other, but only weakly associated with vessel-specific vascular involvement errors, supporting the use of a multi-metric benchmark.}
    \label{fig:metric_correlation}
\end{figure*}

\subsection{Additional method details} 
This section provides additional implementation details.

\subsubsection*{\textbf{TwinTrack} - University of Strasbourg, France}
TwinTrack used a two-stage nnU-Net cascade trained on PANORAMA Batch 4 \cite{kirscher_twintrack_2026}. The second stage relied on a deep ensemble \cite{lakshminarayanan_simple_2017} (typically $M=3$) and used a compound Dice and cross-entropy loss with additional weighting for the tumor class. Post-hoc quantile-based histogram binning \cite{naeini_obtaining_2015} with isotonic regression (PAVA) \cite{barlow_statistical_1973} was applied to calibrate lesion probabilities against mean human agreement.

\subsubsection*{\textbf{CorpuSeg} - Shenzhen Institute of Advanced Technology}
CorpuSeg employed a low-resolution nnU-Net for coarse localization followed by five second-stage full-resolution models, each trained on a different expert annotation. Final predictions were obtained by voxel-wise averaging of the five probabilistic outputs. This design made CorpuSeg one of the clearest examples of prediction-level fusion of multi-rater supervision.

\subsubsection*{\textbf{BreizhSeg} - University of Rennes 1}
BreizhSeg used a pretrained ResidualEncoderUNet initialized from PANORAMA Batch 4 and MSD pancreas tumor data. 
The deterministic model was converted into an Adaptable Bayesian Neural Network \cite{franchi_make_2024} by replacing normalization layers with Bayesian Normalization Layers. 
At inference, five stochastic forward passes were used to approximate the posterior predictive distribution \cite{hemon_towards_2025}, combined later with STAPLE.

\subsubsection*{\textbf{MIC DKFZ} - German Cancer Research Center}
MIC DKFZ pretrained an nnU-Net ResEnc-L model on PANORAMA Batch 4 and fine-tuned an ensemble of five models on CURVAS-PDACVI, using all five annotations together with the STAPLE label as image--ground-truth pairs. Temperature scaling was applied before ensembling. Due to computational constraints, inference was performed without overlapping patches and without test-time augmentation.

\subsubsection*{\textbf{ROISeg} - Chinese Academy of Sciences}
ROISeg trained a full-resolution nnU-Net directly on STAPLE consensus labels. By collapsing the multi-rater annotations into a single fused target before training, the method avoided multi-stage processing and ensembling, at the cost of discarding explicit information about inter-rater disagreement.

\subsubsection*{\textbf{OrdSTAPLE} - Universitat Pompeu Fabra, Sycai Medical}
OrdSTAPLE followed a two-stage curriculum. A first-stage nnU-Net pretrained on PANORAMA Batch 4 was adapted to predict five ordinal agreement levels, and trained with a composite loss including a Rank Probability Score term. In parallel, a second binary model was trained on STAPLE labels. Final probabilistic maps were obtained by averaging both models, followed by thresholding for binary output. In the submitted version, the ensemble combined three ordinal folds and one STAPLE fold, followed by custom Gaussian smoothing.

\subsection{Additional challenge result details}

\subsubsection{Metric correlation analysis}
Figure~\ref{fig:metric_correlation} summarizes the relationships among the benchmark metrics. 
Overlap-based measures (DSC and Thr-DSC) were strongly correlated with each other, whereas their association with vessel-specific vascular involvement errors was markedly weaker. 
This supports the use of a multi-metric benchmark, as strong global volumetric agreement does not necessarily imply accurate assessment of tumor-vessel interfaces.

\begin{figure*}[!t]
    \centering
    \includegraphics[width=\textwidth]{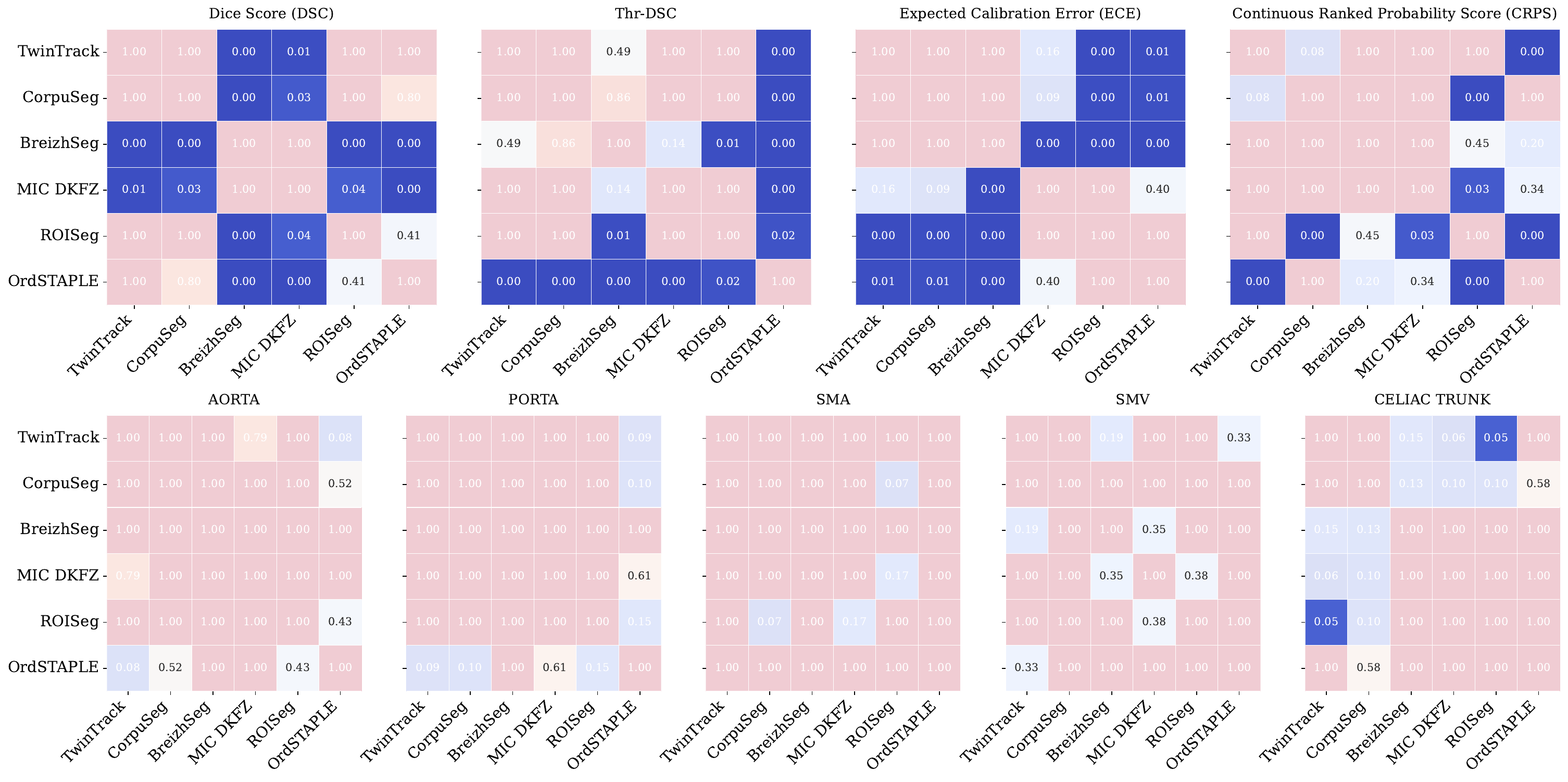}    
    \caption{Pairwise Wilcoxon signed-rank test matrices across benchmark metrics. Heatmaps display p-values for pairwise method comparisons over global segmentation, calibration, and vessel-specific vascular involvement metrics. Several differences remain significant for overlap and calibration, whereas most vascular comparisons do not reach significance, indicating that case-wise variance in vascular assessment is strongly influenced by a subset of highly ambiguous studies shared across methods.}
    \label{fig:wilcoxon_significance}
\end{figure*}
\subsubsection{Pairwise statistical significance}
Pairwise Wilcoxon signed-rank tests are summarized in Fig.~\ref{fig:wilcoxon_significance}. 
For global metrics, several pairwise differences remained significant, particularly in calibration and overlap. 
In contrast, most pairwise comparisons for vessel-specific vascular involvement metrics did not reach significance, indicating that case-wise variance in these structures is strongly driven by a subset of highly ambiguous studies shared across methods.

\begin{figure*}[!t]
    \centering
    \includegraphics[width=\textwidth]{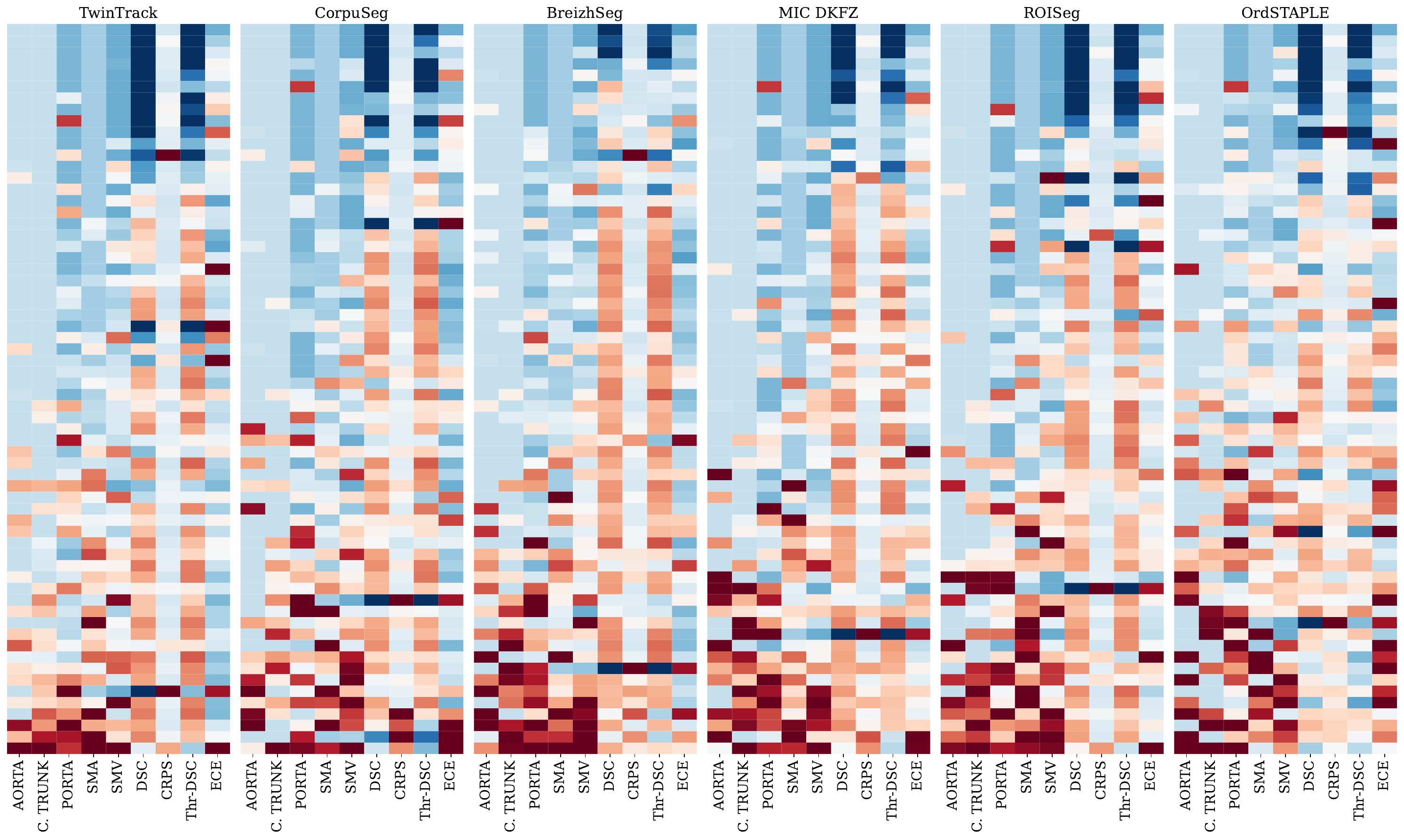} 
    \caption{Case-wise performance signatures across the test cohort. Standardized metric profiles reveal both method-specific trade-offs across multiple architectures.}
    \label{fig:supp_team_performance_signatures}
\end{figure*}
\subsubsection{Case-wise performance signatures}

Figure~\ref{fig:supp_team_performance_signatures} summarizes standardized case-wise performance signatures for each team across the full test cohort. To quantify how much of this variation is driven by the cases themselves rather than by the methods, we decomposed the variance of each metric over the $64 \times 6$ case--method results, matching cases by study identifier. Case identity explained 52--80\% of the variance in every metric, whereas the choice of method explained only 1--5\%. Consistently, per-case results were strongly correlated between teams (mean pairwise Spearman $\rho$ = 0.72 for DSC, 0.78 for Thr-DSC and 0.55--0.95 for the vessel-specific VI errors), indicating that the same studies were difficult for all methods.

Overlap difficulty was closely linked to expert disagreement: each method's DSC correlated with the mean pairwise inter-rater DSC of the case (Spearman $\rho$ = 0.55--0.77; 0.63--0.89 for Thr-DSC). Ten test cases obtained a DSC below 30\% for at least four of the six methods, and four did so for all six. These cases had markedly lower expert agreement (median 30.2\% vs 63.7\% for the full test set) and were mostly small tumours (median expert volume 5.9 vs 17.9~cm$^3$; 8 of 10 below the median).

In contrast, VI errors were essentially unrelated to expert agreement ($|\rho| \leq 0.27$ for all methods and vessels). Within each method, cases with higher DSC tended to have larger, not smaller, VI errors (Spearman $\rho$ = 0.16--0.54 between DSC and the mean W$_1$ across vessels). This reflects case composition rather than a trade-off: small tumours, which are hard to delineate, rarely contact the major vessels and therefore obtain low W$_1$ (Table~\ref{tab:supp_vi_involved}), whereas larger tumours are easier to segment but involve more vessels. Overlap and VI therefore capture different sources of difficulty, which supports reporting both.


\begin{table*}
\centering
\caption{Predicted tumour volume on the 64 test cases. Probabilistic volume is the sum of voxel probabilities; binary volume uses the submitted segmentation. Empty: no foreground voxels; near-empty: binary volume $<1$~cm$^3$.}
\label{tab:supp_volumes}
\small
\begin{tabular}{lcccc}
\toprule
 & Prob. volume (~cm$^3$) & Binary volume (~cm$^3$) & Empty & Near-empty \\
Method & median [IQR] & median [IQR] & (\%) & (\%) \\
\midrule
TwinTrack & 13.0 [4.0--25.7]  & 10.5 [2.9--24.2]  & 18.8 & 18.8 \\
CorpuSeg  & 11.0 [4.8--30.9]  & 10.6 [4.7--31.2]  & 6.2  & 12.5 \\
BreizhSeg & 19.7 [12.4--42.4] & 18.7 [10.6--45.0] & 1.6  & 1.6  \\
MIC DKFZ  & 20.7 [9.1--36.8]  & 20.0 [7.3--36.1]  & 7.8  & 10.9 \\
ROISeg    & 13.4 [6.9--28.0]  & 13.5 [6.9--28.2]  & 1.6  & 6.2  \\
OrdSTAPLE & 25.8 [15.5--40.0] & 18.4 [8.8--33.2]  & 6.2  & 12.5 \\
\midrule
Experts (mean of 5) & -- & 17.9 [9.1--30.3] & 0.0 & 0.0 \\
\bottomrule
\end{tabular}
\end{table*}

\subsubsection{Predicted volumes and empty predictions analysis}

Table~\ref{tab:supp_volumes} reports the predicted tumour volume of each method, computed from the probabilistic map and from the binary segmentation, together with the proportion of empty and near-empty ($<1$ ~cm$^3$) predictions. Median volumes were close to the expert median, but the frequency of missed tumours differed markedly between methods, with TwinTrack producing an empty mask in 18.8\% of test cases. Near-empty predictions occurred mostly in small tumours (13 of the 17 affected cases were below the median expert volume of 17.9 ~cm$^3$).

\subsubsection{VI-error by expert-marked involvement}

Table~\ref{tab:supp_vi_involved} splits the VI error by whether at least one expert marked the vessel as involved. Errors on involved vessels (42–60° on average) dominate the overall VI scores, whereas errors on non-involved vessels, which reflect false-positive contact, are small for all methods except OrdSTAPLE and BreizhSeg.



\input{supp_table_vi_error}


\end{document}

%% file: supp_table_vi_error.tex
\begin{table*}
\centering
\caption{Vascular invasion error (mean $W_1$) on the test set, split by whether at least one expert marked the vessel as involved (Inv.) or not (Not). Number of cases with the vessel involved: PORTA 40, SMV 47, AORTA 17, CELIAC TRUNK 19, SMA 35 (of 64).}
\label{tab:supp_vi_involved}
\small
\begin{tabular}{l*{12}{c}}
\toprule
& \multicolumn{2}{c}{PORTA} & \multicolumn{2}{c}{AORTA} & \multicolumn{2}{c}{SMA} & \multicolumn{2}{c}{SMV} & \multicolumn{2}{c}{CELIAC TRUNK} & \multicolumn{2}{c}{All vessels} \\
\cmidrule(lr){2-3}\cmidrule(lr){4-5}\cmidrule(lr){6-7}\cmidrule(lr){8-9}\cmidrule(lr){10-11}\cmidrule(lr){12-13}
Method & Inv. & Not & Inv. & Not & Inv. & Not & Inv. & Not & Inv. & Not & Inv. & Not \\
\midrule
TwinTrack & 43.9 & 4.2  & 15.9 & 2.5 & 51.9 & 0.7  & 41.7 & 12.9 & 48.1 & 0.3 & 42.5 & 2.9 \\
CorpuSeg  & 45.9 & 1.0  & 25.2 & 0.6 & 51.3 & 0.0  & 52.6 & 5.4  & 49.9 & 0.0 & 47.4 & 0.9 \\
BreizhSeg & 53.7 & 6.3  & 25.4 & 3.2 & 52.0 & 2.0  & 50.4 & 23.9 & 75.4 & 0.0 & 51.9 & 4.8 \\
MIC DKFZ  & 53.1 & 1.0  & 37.3 & 1.1 & 52.1 & 0.0  & 44.9 & 8.5  & 75.7 & 0.0 & 51.5 & 1.4 \\
ROISeg    & 53.1 & 0.0  & 27.2 & 0.9 & 76.7 & 0.0  & 60.0 & 4.3  & 76.8 & 0.0 & 60.4 & 0.7 \\
OrdSTAPLE & 56.3 & 17.7 & 37.2 & 4.9 & 51.3 & 12.4 & 49.6 & 29.2 & 61.1 & 2.0 & 51.7 & 9.9 \\
\midrule
Empty mask & 107.4 & 0.0 & 51.9 & 0.0 & 115.9 & 0.0 & 130.6 & 0.0 & 162.0 & 0.0 & 116.8 & 0.0 \\
\bottomrule
\end{tabular}
\end{table*}